\documentclass[10pt, journal]{IEEEtran}
\usepackage{graphics}
\usepackage{graphicx}
\usepackage{multirow}
\usepackage{times}
\usepackage{adjustbox}
\usepackage{amsmath}
\usepackage{amssymb}
\usepackage{array}
\usepackage{color}
\usepackage{subcaption}
\usepackage{mathtools}
\usepackage{url}
\usepackage{bm}
\usepackage{breqn}
\usepackage{chronosys}

\begin{document}
\title{Transformer-based Generative Adversarial Networks in Computer Vision: A Comprehensive Survey}

\author{
Shiv Ram Dubey, \IEEEmembership{Senior Member,~IEEE}, Satish Kumar Singh, \IEEEmembership{Senior Member,~IEEE}
\thanks{S.R. Dubey and S.K. Singh are with the Computer Vision and Biometrics Laboratory, Indian Institute of Information Technology, Allahabad, Prayagraj, Uttar Pradesh-211015, India (e-mail: srdubey@iiita.ac.in, sk.singh@iiita.ac.in). }
}

\markboth{Transformer-based GANs in Computer Vision: A Comprehensive Survey}
 {Dubey \MakeLowercase{\textit{}}: Bare Demo of IEEEtran.cls for Journals}

\maketitle

\begin{abstract}
Generative Adversarial Networks (GANs) have been very successful for synthesizing the images in a given dataset. The artificially generated images by GANs are very realistic. The GANs have shown potential usability in several computer vision applications, including image generation, image-to-image translation, video synthesis, and others. Conventionally, the generator network is the backbone of GANs, which generates the samples and the discriminator network is used to facilitate the training of the generator network. The discriminator network is usually a Convolutional Neural Network (CNN). Whereas, the generator network is usually either an Up-CNN for image generation or an Encoder-Decoder network for image-to-image translation. The convolution-based networks exploit the local relationship in a layer, which requires the deep networks to extract the abstract features. Hence, CNNs suffer to exploit the global relationship in the feature space. However, recently developed Transformer networks are able to exploit the global relationship at every layer. The Transformer networks have shown tremendous performance improvement for several problems in computer vision. Motivated from the success of Transformer networks and GANs, recent works have tried to exploit the Transformers in GAN framework for the image/video synthesis. This paper presents a comprehensive survey on the developments and advancements in GANs utilizing the Transformer networks for computer vision applications. The performance comparison for several applications on benchmark datasets is also performed and analyzed. The conducted survey will be very useful to deep learning and computer vision community to understand the research trends \& gaps related with Transformer-based GANs and to develop the advanced GAN architectures by exploiting the global and local relationships for different applications. 
\end{abstract}


\section{Introduction}\label{Introduction}
\begin{figure}
    \centering
    \includegraphics[width=0.8\columnwidth]{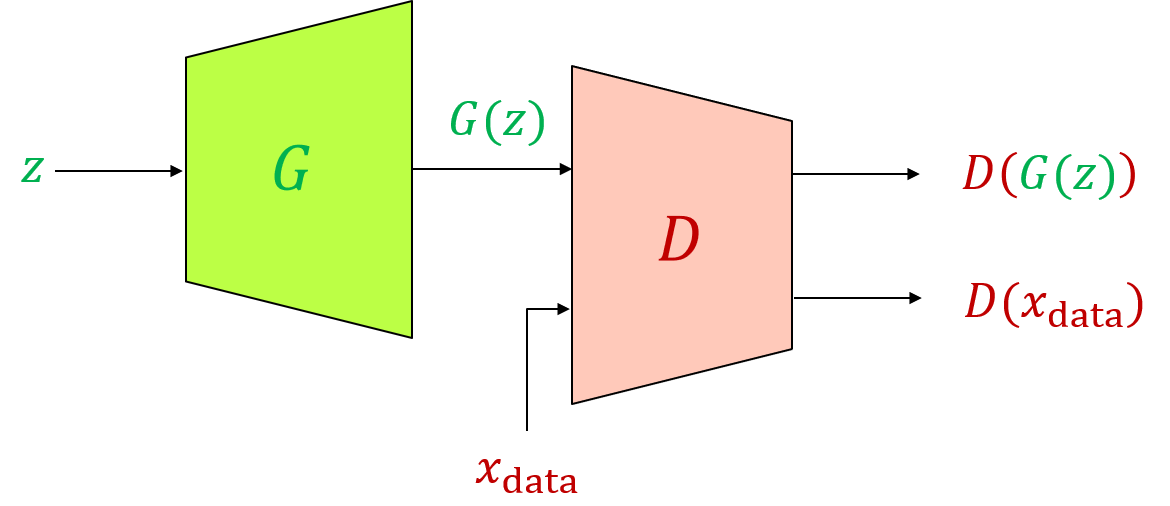}
    \caption{A conceptual representation of GAN \cite{gan}.}
    \label{fig:gan_conceptual}
\end{figure}

\IEEEPARstart{G}{enerative} Adversarial Network (GAN) was introduced by Goodfellow et al. \cite{gan} in 2014 to synthesize images in a given distribution. GAN consists of two neural networks, namely Generator ($G$) and Discriminator ($D$) as shown in Fig. \ref{fig:gan_conceptual}. These networks are trained jointly in an adversarial manner. The generator network outputs a synthesized image from a random latent vector as input. Whereas, the discriminator network classifies the generated images in fake category and actual images in real category. Since its inception, several variants of GANs have been introduced to synthesize high quality images, such as Deep Convolutional GAN (DCGAN) \cite{dcgan}, Wasserstein GAN (WGAN) \cite{wgan}, Least Square GAN (LSGAN) \cite{lsgan}, ProgessiveGAN \cite{progressivegan}, StyleGAN \cite{staylegan}, \cite{karras2020analyzing}, DR-GAN \cite{tan2022dr}, DHI-GAN \cite{lin2022dhi}, ZeRGAN \cite{diao2022zergan}, Tensorizing GAN \cite{yu2021tensorizing}, etc.
Image-to-image translation is one of the major applications of GANs. The early work includes Conditional GAN \cite{conditional_gan} based Pix2pix model \cite{pix2pix} which takes images in a given domain as input and generates images in the target domain as output. In order to do so, the Pix2pix model modifies the generator network of GAN and uses the Encoder-Decoder framework, where the Encoder network is a CNN and the Decoder network is an Up-CNN. Pix2pix makes use of a pixel error to train the model along with the adversarial training, hence requires the paired dataset. To overcome this issue, CycleGAN \cite{cyclegan} uses a cyclic consistency loss that works on the unpaired datasets. Several variants of GAN have been proposed for image-to-image translation, such as PCSGAN \cite{pcsgan}, MUNIT \cite{munit}, CUT \cite{CUT}, CouncilGAN \cite{councilgan}, MobileAR-GAN \cite{mobileargan}, etc.

A huge progress in GAN models, its objective functions and training strategies has been observed, such as an overview on the working of GAN model and its variants is presented in \cite{hong2019generative}, the challenges in GAN models with possible solutions are summarized in \cite{saxena2021generative}, a survey on GAN's variants, applications, and training is performed in \cite{jabbar2021survey}, and a review on GAN's algorithms, theory, and applications in conducted by Gui et al. \cite{gui2021review}.
The GAN-based models have been very successful for several applications as indicated by different surveys/reviews on GANs, such as computer vision \cite{wang2021generative}, image-to-image translation \cite{pang2021image}, face generation \cite{toshpulatov2021generative}, \cite{kammoun2022generative}, medical image analysis \cite{alamir2022role}, \cite{yi2019generative}, video generation \cite{aldausari2022video}, spatio-temporal data analysis \cite{gao2022generative}, and text-to-image synthesis \cite{zhou2021survey}.

In recent years, Transformer networks \cite{vaswani2017attention} have received lots of attention due to its outstanding performance. Transformer network is based on multi-head self-attention modules, which capture the global relationship in the data while transforming the input feature to output feature. Motivated from the success of Transformer networks, several variants have been introduced, such as Bidirectional Transformers (BERT) \cite{bert}, Generative Pre-Training Transformer (GPT) \cite{gpt3}, Swin Transformer \cite{swin_transformer}, Vision Transformer (ViT) \cite{vit}, Multitask ViT \cite{tian2023end}, Tokens-to-token ViT \cite{token_vit}, Video Transformer \cite{alfasly2022effective}, HSI Denoising Transformer (Hider) \cite{chen2022hider}, Word2Pix Transformer \cite{zhao2022word2pix}, and many more.
Transformer networks have been very successful for different problems of natural language processing and computer vision, as indicated by several survey papers. A survey of efficient transformer models is presented in \cite{tay2022efficient}. A review of BERT-based transformer models for text-based emotion detection is conducted by Acheampong et al.  \cite{acheampong2021transformer}. Kalyan et al. present a survey of transformer-based biomedical pretrained language models \cite{kalyan2021ammu}. The potential of transformers is also witnessed in vision \cite{khan2022transformers}, \cite{han2022survey}. Shin et al. highlights the transformer architectures for cross-modal vision and language tasks in terms of different perspectives and prospects \cite{shin2022perspectives}.

Motivated from the wider acceptance of GAN and Transformer models, recently, some researchers have proposed the Transformer-based GAN models for image generative tasks \cite{transgan}, \cite{transformer_attngan}, \cite{swingan}, \cite{vitgan}, \cite{uvcgan}. The recently developed Transformer-based GAN models have shown very promising performance for different applications of image and video synthesis. The contributions of this paper are as follows:
\begin{itemize}
    \item As per our best knowledge, no survey paper exists on GAN models utilizing Transformer networks. Hence, this paper provides a comprehensive survey on the developments in Transformer-based GANs for computer vision applications.
    \item The categorization of models is performed for different image and video applications, such as image generation, image translation, image inpainting, image restoration, image reconstruction, image enhancement, image super-resolution, image colorization, video generation, video inpainting, video translation, etc.
    \item The comparison of models is also performed in terms of the generator architecture, discriminator architecture, loss functions and datasets used for different image and video applications.
    \item The experimental results comparison and analysis is also conducted for different applications using standard metrics on benchmark datasets to provide the status of state-of-the-art Transformer-based GANs.
    \item The research trends and potential future directions are also narrated with a very concise view to benefit the researchers working on deep learning and computer vision problems.
\end{itemize}

Section \ref{background} presents a brief of Generative Adversarial Network and Transformer Network. Sections III-VI are devoted to transformer-based GAN models for image generation, image-to-image translation, video applications and miscellaneous applications, respectively. Section VII provides the conclusion and future directions.

\section{Background} \label{background}
In this section, the background of GANs and Transformers is presented. 

\subsection{Generative Adversarial Network}
Generative Adversarial Network (GAN) was introduced by Goodfellow et al. \cite{gan} in 2014 for image generation. A conceptual representation of GAN is depicted in Fig. \ref{fig:gan_conceptual}. The generator network ($G$) in GAN synthesizes a new sample ($G(z)$) from a random latent vector/noise ($z$), i.e., $G: z \rightarrow G(z)$, where $z \in \mathbb{R}^d$ is sampled from a uniform probability distribution $p_v$ and $d$ is the dimensionality of $z$. If the probability distribution of the generated samples ($G(z)$) and real samples ($x_{data}$) are given by $p_{model}$ and $p_{data}$, respectively, then we want to learn $p_{model}$ that matches $p_{data}$. The training of the generator network is facilitated by the discriminator network ($D$). The purpose of the discriminator network is to distinguish the generated samples from the real samples. The output of the discriminator network is considered as the probability of real samples. Hence, the discriminator network tries to produce $D(x_{data}) \approx 1$, $\forall x_{data} \sim{p_{data}}$ and $D(G(z)) \approx 0$, $\forall z \sim{p_v}$, where $p_{data}$ represents the probability distribution of real samples. However, at the same time, the generator network tries to fool the discriminator network and to achieve $D(G(z)) \approx 1$, $\forall z \sim{p_v}$. It makes training of the GAN model as a min-max optimization. The objective function of GAN is given as,
\begin{equation}
\begin{aligned}
    {\mathcal{L}}_{GAN}(G, D) = 
    & Min_{G}Max_{D}(  \mathbb{E}_{x_{data}\sim{p_{data}}} [\log D(x_{data})] + \\
    & \mathbb{E}_{z\sim{p_v}}[\log(1-D(G(z)))] )
\end{aligned}
\end{equation}
where ${\mathcal{L}}_{GAN}(G, D)$ is the adversarial loss function.

\begin{figure}
    \centering
    \includegraphics[width=\columnwidth]{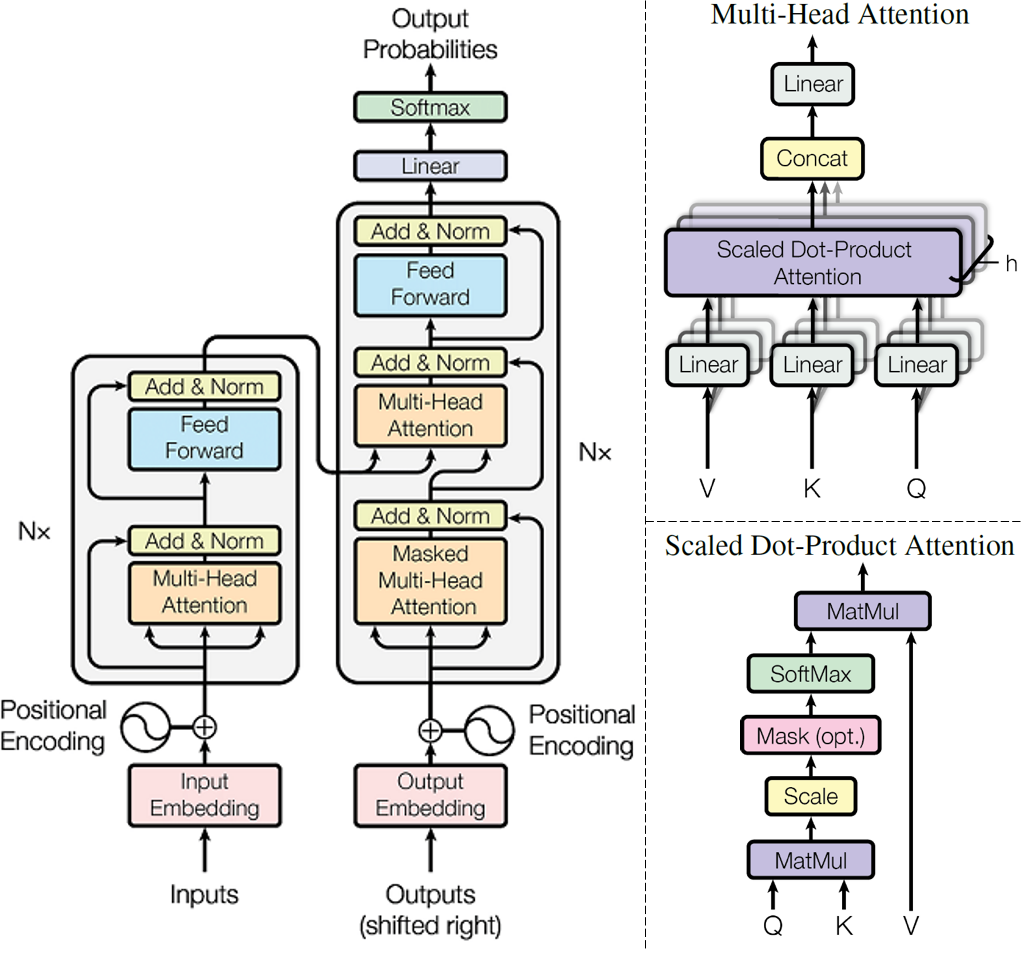}
    \caption{(\textit{left}) Transformer network, (\textit{right, top}) Multi-head attention mechanism, and (\textit{right, bottom}) Scaled dot-product attention module \cite{vaswani2017attention}.}
    \label{fig:transformer}
\end{figure}

Considering unlimited training data and unlimited capacity for generator and discriminator:
\begin{itemize}
    \item The objective ${\mathcal{L}}_{GAN}(G, D)$ is equivalent to the Jensen-Shannon divergence between $p_{data}$ and $p_{model}$ and global optimum (Nash equilibrium) is given by $p_{data}$ and $p_{model}$.
    \item If at each step, $D$ is allowed to reach its optimum given $G$, and $G$ is updated to decrease ${\mathcal{L}}_{GAN}(G, D)$, then $p_{model}$ will eventually converge to $p_{data}$.
\end{itemize}

The discriminator network is usually a Convolutional Neural Network (CNN) and works as a binary classifier. The last layer of discriminator network is a Sigmoid function which produces the output as the probability for real class. The discriminator network is only required for training purpose. The generator network is usually an Up-CNN which can take the input having low dimensionality and produce the output with high dimensionality. The generator network architecture is suitably modified for different type of data and applications, such as encoder-decoder based networks are used for image-to-image translation \cite{pang2021image} and RNN-CNN based networks are used for text-to-image synthesis \cite{zhou2021survey}.

\subsection{Transformer Network}
The Transformer network utilizes a self-attention mechanism introduced by Vaswani et al. in 2017 \cite{vaswani2017attention}. Originally, the Transformer was proposed for a machine translation task where the Encoder and Decoder networks are built using Transformers as shown in Fig. \ref{fig:transformer} (\textit{left}). First, the feature embedding from input is computed and combined with the positional embedding to generate the input feature embedding of $k$ dimensionality for Transformer. A Transformer block in Encoder consists of a multi-head attention module followed by a feed forward module having skip connection and normalization. However, a Transformer block in Decoder includes an additional masked multi-head attention module. Basically, a Transformer block transforms an input feature embedding ($u \in \mathbb{R}^k$) into an output feature embedding ($v \in \mathbb{R}^k$). The Transformer block is repeated $N\times$ in Encoder and Decoder networks. The multi-head attention module is illustrated in Fig. \ref{fig:transformer} (\textit{right, top}). It is basically the concatenation of the output of several independent scaled dot-product attention followed by a linear layer. The use of multiple attention heads facilitates the extraction of features of different important regions or characteristics. The scaled dot-product attention mechanism is depicted in Fig. \ref{fig:transformer} (\textit{right, bottom}). The input ($u$) to a Transformer block is projected into the Query ($Q$), Key ($K$), and Value ($V$) using linear layers having weights $W_Q$, $W_K$, and $W_V$, respectively. The output of the scaled dot-product attention module is computed as,
\begin{equation}
\text{Attention}(Q,K,V) = \text{softmax}(\frac{QK^T}{\sqrt{d_k}})V
\end{equation}
where $d_k$ is the dimensionality of $Q$ and $K$. The masking in scaled dot-product attention module is used only in Decoder network. The feed forward module consists of a linear layer followed by the ReLU activation function followed by another linear layer. 

\begin{figure}
    \centering
    \includegraphics[width=\columnwidth]{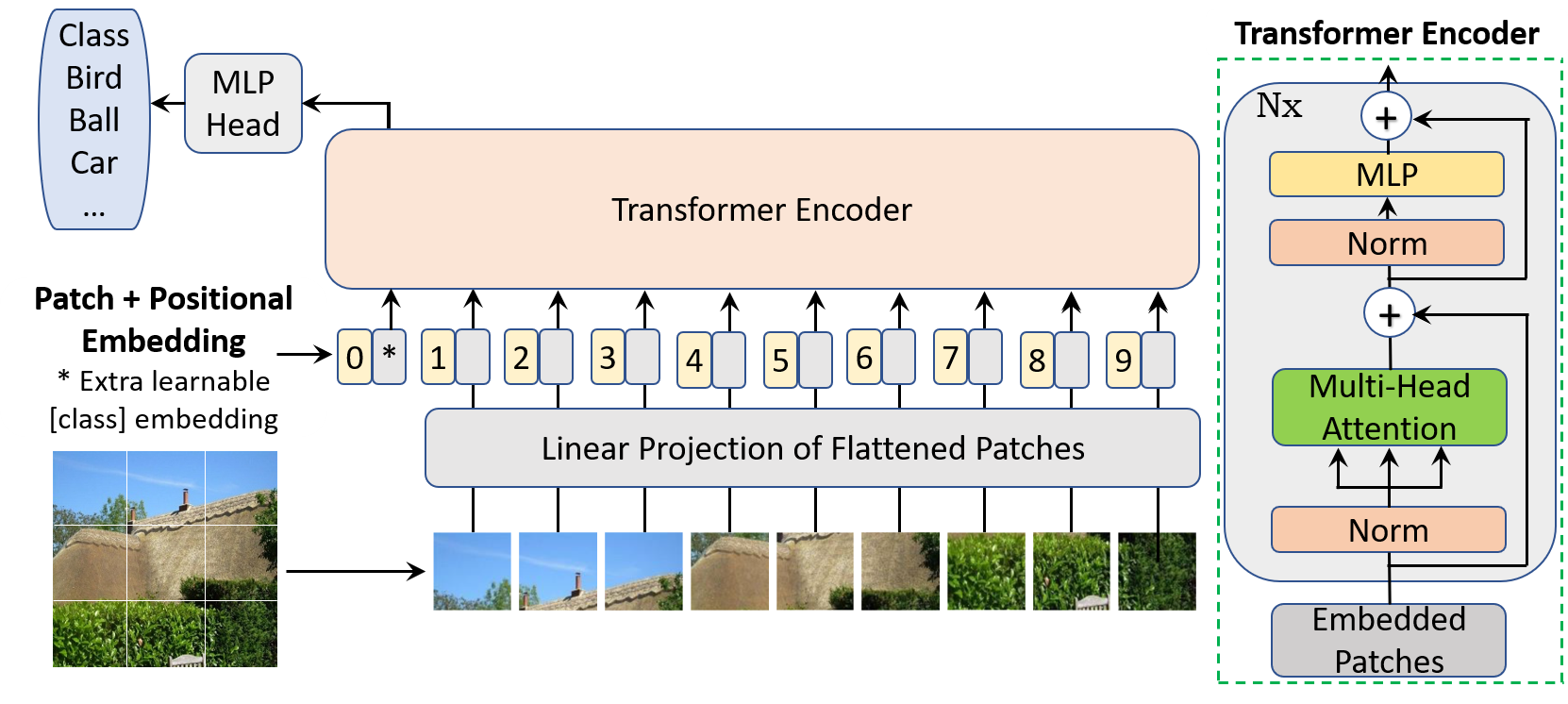}
    \caption{Vision Transformer (ViT) for image classification \cite{vit}.}
    \label{fig:vit}
\end{figure}

Vision Transformer (ViT) \cite{vit} is a variant of the transformer network which facilitates to utilize it for the image input. Basically, it divides the images into patches. The feature embeddings extracted from these patches are used as the input to Transformer network as depicted in Fig. \ref{fig:vit}. Transformer Encoder consists of $N$ Transformer blocks. An additional token for class embedding is added in ViT which is used as the input to the Multi-layer Perceptron (MLP) head to generate the class scores for different classes. Though Transformer and ViT networks were originally proposed for the machine translation and image classification tasks, respectively, they are very heavily utilized for different problems of computer vision in recent years. In this paper, we provide the advancements in GAN models using Transformer networks for image and video synthesis and analyze from different perspectives.

\begin{table*}[!t]
\caption{A summary of Transformer-based GANs for image generation.}
\centering
\begin{tabular}{p{0.23\columnwidth}|p{0.15\columnwidth}|p{0.4\columnwidth}|p{0.32\columnwidth}|p{0.34\columnwidth}|p{0.32\columnwidth}}
    \hline
    \textbf{Model} & \textbf{Venue} & \textbf{Generator} & \textbf{Discriminator} & \textbf{Objective Function} & \textbf{Datasets} \\
    \hline
    GANsformer \cite{gansformer} & ICML'21 & Bipartite Transformer having simplex and duplex attention & Attention CNN-based discriminator  & Loss functions of StyleGAN & CLEVR \\
    \hline
    GANformer2 \cite{ganformer2} & NeurIPS'21 & Generator works in two stages: layout generation and layout to scene translation & One CNN for real vs. fake and one U-Net for semantic matching & Adversarial loss, Semantic-matching loss and Segment-fidelity loss & CLEVR, Bedrooms, CelebA, Cityscapes and COCO\\
    \hline
    TransGAN \cite{transgan} & NeurIPS'21 & A Transformer-based generator that progressively increases feature resolution & Transformer-based discriminator that takes input at multiple scales & WGAN-GP loss & CIFAR-10, STL-10, CelebA, CelebA-HQ and LSUN Church \\
    \hline
    HiT \cite{hit} & NeurIPS'21 & Multi-Axis Nested Transformer at low-resolution and Implicit Functions at high-resolution & ResNet-based discriminator & Non-saturating logistic GAN loss, R1 gradient penalty to only discriminator & ImageNet, CelebA-HQ and FFHQ\\
    \hline
    TokenGAN \cite{tokengan} & NeurIPS'21 & Visual Transformer with content and style tokens & Discriminator of StyleGAN2 & Non-saturating logistic adversarial loss, R1 regularization to only discriminator & FFHQ and LSUN Church \\
    \hline
    VQGAN \cite{vqgan} & CVPR'21 & CNN-based image constituents vocabulary and Transformer-based modeling of vocabulary composition within high-resolution image & CNN-based discriminator & Adversarial loss, reconstruction loss, commitment loss and perceptual reconstruction loss & ImageNet, ImageNet-Animal, LSUN Churches \& Towers, COCO-Stuff, ADE20K, CelebA-HQ and FFHQ \\
    \hline
    Styleformer \cite{styleformer} & CVPR'22 & Transformer with Styleformer Encoders having Increased Multi-Head Self-Attention & Discriminator of StyleGAN2-ADA & Losses of StyleGAN2-ADA &  CIFAR-10, STL-10, CelebA, LSUN-Church, CLEVR and Cityscapes \\
    \hline
    StyleSwin \cite{styleswin} & CVPR'22 & Style-based GAN with Transformer having double attention modules & Wavelet-based discriminator &  Non-saturating GAN loss with R1 gradient penalty and spectral normalization on the discriminator &  FFHQ, CelebA-HQ and LSUN Church\\
    \hline
    ViTGAN \cite{vitgan} & ICLR'22 & ViT-based ordered patch generator & ViT-based discriminator & Non-saturating logistic adversarial loss & CIFAR-10, CelebA and LSUN Bedroom \\
    \hline
    Unleashing Transformer \cite{unleashing_transformer} & ECCV'22 & Trained Transformer using Masked Vector-Quantized tokens prediction & Traditional discriminator & Vector-Quantized loss, generator loss and reconstruction loss & FFHQ, LSUN Bedroom and LSUN Churches \\ \hline
    Swin-GAN \cite{swin_gan_tvc} & TVC'22 & Swin Transformer-based generator & Swin Transformer-based multi-scale discriminator & WGAN-GP loss & CIFAR-10 and Anime images \\ \hline
    PGTCEGAN \cite{pgtcegan} & SMC'22 & Capsule Embedding based Progressive Growing Transformer & CNN with multi-scale input in different layer & WGAN-GP loss & CIFAR-10, CelebA and LSUN-Church  \\ \hline
    MedViTGAN \cite{medvitgan} & ICPR'22 & ViT Encoder-based generator & ViT Encoder-based discriminator in conditional GAN setup & WGAN-GP loss with adaptive hybrid loss weighting mechanism & Histopathology image dataset: PatchCamelyon (PCam) and BreakHis \\ \hline
    PTNet3D \cite{ptnet3d} & IEEE-TMI'22 & U-shape generator with performer encoder, transformer bottleneck and performer decoder & 3D ResNet-18 model pretrained on Kinetics-400 dataset & Adversarial loss, Perceptual loss and Mean square error & MRI datasets: Developing Human Connectome Project (dHCP) and longitudinal Baby Connectome Project (BCP) \\ \hline
    SLATER \cite{slater} & IEEE-TMI'22 & Generator uses cross-attention transformers with input from a mapper & CNN-based discriminator & Non-saturating logistic adversarial loss, gradient penalty for discriminator & MRI synthesis: brain MRI data from fastMRI\\ \hline
    SwinGAN \cite{swingan} & CBM'23 & Swin Transformer U-Net-based frequency-domain and image-domain generators & CNN-based discriminator & Adversarial loss, k-space loss and image domain loss & MRI reconstruction: IXI brain dataset \\ \hline
    3D Face Transformer \cite{transformer_3d_face_reconstruction} & IEEE-TCSVT'22 & Residual blocks followed by a multi-layer transformer encoder-based generator & Traditional discriminator & Adversarial loss, L1 loss, Edge loss, L1 loss on the transformer outputs and Self-supervised reprojection consistency loss & 3D Face reconstruction: 300W-LP, AFLW, AFLW2000-3D, NoW, In-the-wild images \\ \hline
\end{tabular}
\label{tab:image_generation}
\end{table*}

\section{Transformer-based GANs for Image Generation}
The image generation has been very important application of GANs. Several improvements in GAN have been validated for this task. The researchers have also exploited the Transformer-based GANs for image generation of different types, such as objects, scenes, medical, etc. A summary of the different models is presented in Table \ref{tab:image_generation} in terms of the generator, discriminator, losses and datasets.

\subsection{Image Generation}
In the initial attempts, Transformers are utilized in an auto-regressive manner \cite{gpp}, \cite{vqgan} for the image generation. However, these methods are not very efficient in terms of the inference speed.
Jiang et al. in 2021 conducted the initial study on a pure Transformer-based GAN, dubbed TransGAN \cite{transgan}, which is completely free of convolution operations. TransGAN contains a Transformer-based generator network that increases the resolution of features in a progressive manner and a Transformer-based discriminator network. Both the generator and discriminator networks utilize the grid-transformer blocks. TransGAN achieve Fréchet Inception Distance (FID) of 9.26 on CIFAR-10 dataset \cite{cifar}.
Hudson and Zitnick \cite{gansformer} proposed generative adversarial transformer (GANsformer) for image generation. GANsformer makes use of the bipartite transformer which is a stack of alternative simplex or duplex attention modules. The results of GANsformer are demonstrated on four benchmark datasets. GANsformer is further extended to GANformer2 by incorporating explicit and strong structural priors \cite{ganformer2}.
Researchers have experienced that training GANs with Transformers is challenging. Xu et al. \cite{stransg} proposed STrans-G generator network and STrans-D discriminator network for image generation by studying the characteristics of Transformer in GAN framework. STrans-G is CNN free network. It is noticed in \cite{stransg} that the residual connections in self-attention layers is not good for Transformer-based discriminators and conditional generators. Zeng et al. \cite{tokengan} proposed a TokenGAN which exploits a token-based transformer generator to assign the styles to the content tokens through attention mechanism for controlling the image synthesis. TokenGAN generates the high-fidelity images at high-resolution (i.e., 1024 × 1024 size).

Zhao et al. \cite{hit} proposed an efficient high-resolution image transformer (HiT) based GAN model for image generation. HiT exploits the multi-axis blocked self-attention module at low-resolution synthesis and removes the costly global self-attention module at high-resolution synthesis. In an another study, the Transformer-based generator network is utilized with Convolution-based discriminator by Durall et al. \cite{transG_convD} for image generation. The experimental results using the generator of TransGAN and discriminator of SNGAN in \cite{transG_convD} show an improved FID of 8.95 on the CIFAR-10 dataset. Park and Kim \cite{styleformer} introduced the Styleformer model which contains the Transformer structure-based generator network for synthesizing the images using style vectors. An attention style injection module is introduced in Styleformer for the modulation and demodulation of style with self-attention. Styleformer shows an outstanding performance for image generation on CIFAR-10 dataset with an FID score of 2.82. Recently, in 2022, a StyleSwin model is proposed by Zhang et al. \cite{styleswin} for image generation. StyleSwin is a  Swin Transformer-based GAN model in a style-based architecture for high-resolution synthesis. In order to exploit the local and shifted window contexts, StyleSwin works based on a double attention mechanism. The transformer is also exploited as the generator network with conditional GAN for image generation by Xi et al. \cite{transformer_cgan}. An unconstrained Transformer architecture is used as the backbone in \cite{unleashing_transformer} which performs the parallel prediction of Vector-Quantized tokens and achieves the competitive FID scores on benchmark datasets. 

A progressive growing transformer with capsule embedding GAN (PGTCEGAN) is introduced by Jiang et al. \cite{pgtcegan}. The generator network generates the images gradually using the transformer. The capsule encoder is utilized to generate the positional embedding. Wang et al. proposed a Swin-GAN by exploiting the shifted window attention mechanism-based transformer to gradually increase the resolution \cite{swin_gan_tvc}. Basically, the generator and discriminator networks of Swin-GAN use the Swin Transformer Blocks. The discriminator in Swin-GAN receives the input by dividing the image into patches of different sizes. Relative position coding, layer normalization and data enhancement are also exploited by Swin-GAN to improve the stability.

Lee et al. utilized the vision transformer in GAN framework and proposed ViTGAN \cite{vitgan}. It is noticed that the training of ViT discriminator is unstable due to the self-attention module not able to cope with the existing GAN regularization methods. ViTGAN exploits different regularization techniques to tackle this problem, such as the Lipschitzness of Transformer discriminator, improved spectral normalization, overlapping image patches and convolutional projection. By carefully designing the ViT generator, ViTGAN is able to converge with comparable performance to the leading CNN-based GAN models for image generation. It is also observed that by using the generator network of ViTGAN (ViTGAN-G) and a discriminator network of (StyleGAN2-D), the performance of the image generation task is improved. A HybridViT model is introduced in \cite{hybridvit} by integrating the ViT architecture into diffusion denoising probability models. HybridViT model is utilized for joint image generation and classification.

\begin{table}[!t]
    \caption{The results comparison of Transformer-based GANs for image generation in terms of FID score. \textbf{Best} and \textit{second best} results are highlighted in \textbf{bold} and \textit{italic}, respectively, even in other tables also.}
    \centering
    \resizebox{\columnwidth}{!}{%
    \begin{tabular}{p{0.23\columnwidth} p{0.13\columnwidth} p{0.1\columnwidth} p{0.08\columnwidth} p{0.08\columnwidth} p{0.1\columnwidth} p{0.1\columnwidth}}
    \hline
    Method & Venue & CIFAR10 ($32^2$) & STL10 ($48^2$) & CelebA ($64^2$) & ImageNet ($128^2$) \\
    \hline
    TransGAN \cite{transgan} & NeurIPS'21 & 9.26 & \textit{18.28} & 5.01 & - \\
    HiT \cite{hit} & NeurIPS'21 & - & - & - & \textit{30.83}\\
    STrans-G \cite{stransg} & arXiv'21 & \textbf{2.77} & - & \textbf{2.03} & \textbf{12.12} \\
    Styleformer \cite{styleformer} & CVPR'22 & \textit{2.82} & \textbf{15.17} & 3.92 & - \\
    PGTCEGAN \cite{pgtcegan} & SMC'22 & 8.43 & - & \textit{3.59} & - \\
    Swin-GAN \cite{swin_gan_tvc} & TVC'22 & 9.23 & - & - & - \\
    ViTGAN \cite{vitgan} & ICLR'22 & 4.92 & - & 3.74 & - \\
    \hline
    \end{tabular} }
    \label{tab:results_image_generation}
\end{table}

\subsection{Medical Image Synthesis}
The image synthesis in medical domain is a very demanding application of GANs. The synthetic medical images are generally useful for the data augmentation to tackle the requirement of large-scale datasets of deep learning models. Due to the improved quality of generated images, the Transformer-based GANs have also been extensively utilized for medical image synthesis of different modality.

In 2021, Korkmaz et al. \cite{gvtrans} proposed a generative vision transformer-based GVTrans model for synthesizing the MRI data from noise variables and latent space. GVTrans model is tested for unsupervised MRI reconstruction with promising performance. In 2022, Zhang et al. introduced a PTNet3D model \cite{ptnet3d} which uses the pyramid transformer network for the synthesis of 3D high-resolution longitudinal infant brain MRI data. PTNet3D model makes use of attention in transformer and performer layers. The synthesis accuracy of PTNet3D is superior with better generalization capability. The corrupted scan replaced by synthesized image through PTNet3D leads to better infant whole brain segmentation. A multi-contrast multi-scale transformer (MMT) is proposed by Liu et al. \cite{mmt} for MRI missing data imputation. In order to capture inter- and intra-contrast dependencies, MMT exploits the multi-contrast Swin transformer blocks. MMT is tested for image synthesis on two multi-contrast MRI datasets with appealing performance. 

In 2022, Li et al. proposed MedViTGAN for histopathological image synthesis for data augmentation \cite{medvitgan}. MedViTGAN is a vision transformer-based conditional GAN where both the generator and discriminator networks are designed based on the transformer encoder module. The performance of MedViTGAN is highly improved as compared to the CNN models, such as ResNet and DenseNet. Li et al. \cite{hvitgan} further extended the MedViTGAN using an auxiliary discriminator head for classification. Korkmaz et al. \cite{slater} proposed a zero-Shot Learned Adversarial TransformERs (SLATER) for unsupervised MRI reconstruction. SLATER utilizes cross-attention transformers with the deep adversarial network to project latent variables and noise into coil-combined MR images. 

\begin{table}[!t]
    \caption{The results comparison of Transformer-based GANs for high-resolution image generation in terms of FID score. Here, HiT-B refers to HiT with large model capacity.}
    \centering
    \resizebox{\columnwidth}{!}{%
    \begin{tabular}{p{0.235\columnwidth} p{0.13\columnwidth} p{0.07\columnwidth} p{0.075\columnwidth} p{0.07\columnwidth} p{0.07\columnwidth} p{0.08\columnwidth}}
    \hline
    Method & Venue & FFHQ ($256^2$) & CelebA-HQ ($256^2$) & LSUN ($256^2$) & FFHQ ($1024^2$) & CelebA-HQ ($1024^2$)\\
    \hline
    VQGAN \cite{vqgan} & CVPR'21 & 11.40 & 10.70 & - & - & -\\
    TransGAN \cite{transgan} & NeurIPS'21 & - & 9.60 & 8.94 & - & - \\
    GANsformer \cite{gansformer} & ICML'21 & 7.42 & - & 6.51 & - & - \\
    GANformer2 \cite{gansformer} & NeurIPS'21 & 7.77 & - & 6.05 & - & - \\
    TokenGAN \cite{tokengan} & NeurIPS'21 & 5.41 & - & 5.56 & - & - \\
    STrans-G \cite{stransg} & arXiv'21 & 4.84 & - & - & - & - \\
    HiT-B \cite{hit} & NeurIPS'21 & \textit{2.95} & \textit{3.39} & - & \textit{6.37} & \textit{8.83}\\
    PGTCEGAN \cite{pgtcegan} & SMC'22 & - & - & \textit{3.92} & - & - \\
    StyleSwin \cite{styleswin} & CVPR'22 & \textbf{2.81} & \textbf{3.25} & \textbf{2.95} & \textbf{5.07} & \textbf{4.43} \\
    \hline
    \end{tabular}}
    \label{tab:results_hr_image_generation}
\end{table}

\subsection{Results Comparison and Analysis}
The results comparison of Transformer-based GANs is reported in Table \ref{tab:results_image_generation} and Table \ref{tab:results_hr_image_generation}. The Fréchet inception distance (FID) metric is compared to the different models on benchmark datasets. A lower FID score represents the better synthesized image quality. Table \ref{tab:results_image_generation} summarizes the FID scores on CIFAR10 \cite{cifar}, STL10 \cite{stl}, CelebA \cite{celeba} and ImageNet \cite{imagenet} datasets to generate the images of size $32\times32$, $48\times48$, $64\times64$ and $128\times128$, respectively. It is noticed that STrans-G \cite{stransg} provides best FID scores for CIFAR10, CelebA and ImageNet datasets. The Swin architecture at resolution more than $16\times16$ provides local attention in STrans-G which is beneficial for qualitative image generation. The performance of Styleformer is also promising with best on STL10 and second best on CIFAR10 dataset. The FID scores for high-resolution image generation are illustrated in Table \ref{tab:results_hr_image_generation} over Flickr-Faces-HQ (FFHQ) \cite{staylegan}, CelebA-HQ \cite{progressivegan} and LSUN Church \cite{lsun_church} datasets to generate the images of size $256\times256$. The results for generating the images of size $1024\times1024$ are also included over FFHQ and CelebA datasets. It is evident that StyleSwin \cite{styleswin} is the best performing model over all the datasets for high-resolution image synthesis as the capabilities of both StyleGAN and Swin transformer are inherited by StyleSwin. Moreover, the local-global positional encoding in StyleSwin is able to maintain a good trade-off between local and global context while generating the high-resolution images. The results of HiT \cite{hit} and PGTCEGAN \cite{pgtcegan} are also very promising.

\subsection{Summary}
Following is the summary from the survey of Transformer-based GANs for image generation:
\begin{itemize}
    \item The majority of the transformer-based GAN models use the transformer-based generator network and CNN-based discriminator network. However, few models also exploit the transformers in discriminator network.
    \item The uses of both local and global information are imnportant. Hence, several models try to exploit the CNN-based encoder and decoder modules along with transformer module in the generator network.
    \item ViT and Swin transformers are heavily exploited for the generator network. It is observed that ViT and Swin transformer are useful to preserve the local context.
    \item The objective function for different models generally includes Adversarial GAN loss and perceptual loss. Other losses such as L1 loss, gradient penalty and edge loss are also exploited by some models.
    \item The CIFAR10, STL10, CelebA, ImageNet, FFHQ, CelebA-HQ and LSUN Church are the most common benchmark datasets for image generation.
    \item The combination of StyleGAN with Swin transformer becomes very powerful to generate high-resolution images.
    \item The transformer-based GANs are very useful in tackling the data limitation problem in the medical domain by generating very realistic medical images.
\end{itemize}

\begin{table*}[!t]
\caption{A summary of Transformer-based GANs for image-to-image translation.}
\centering
\begin{tabular}{p{0.21\columnwidth}|p{0.15\columnwidth}|p{0.4\columnwidth}|p{0.25\columnwidth}|p{0.36\columnwidth}|p{0.4\columnwidth}}
    \hline
    \textbf{Model} & \textbf{Venue} & \textbf{Generator} & \textbf{Discriminator} & \textbf{Objective Function} & \textbf{Application \& Datasets} \\
    \hline
    InstaFormer \cite{instaformer} & CVPR'22 & Generator with ViT encoder blocks consisting of adaptive instance normalization & Traditional discriminator & Adversarial loss, Global and Instance-level content loss, Image and Style reconstruction loss & Image translation: INIT, Domain adaptation: KITTI to Cityscapes \\ \hline
    UVCGAN \cite{uvcgan} & WACV'23 & UNet-ViT Generator in CycleGAN framework (pre-trained on the image inpainting task) & CycleGAN discriminator with gradient penalty & GAN loss, cycle-consistency loss, and identity-consistency loss & Unpaired image translation: Selfie to Anime, Anime to Selfie, Male to Female, Female to Male, Remove Glasses, and Add Glasses\\ \hline
    SwinIR \cite{swinir} & ICCV'21 & Generator with residual Swin Transformer blocks & Traditional discriminator for super-resolution & Super-resolution: Pixel loss, GAN loss and perceptual loss & Super-resolution: Set5, Set14, BSD100, Urban100, Manga109, RealSRSet \\ \hline
    RFormer \cite{rformer} & IEEE-JBHI'22 & Transformer-based U-shaped generator with window-based self-attention block & Transformer-based discriminator with window-based self-attention block & Adversarial loss, Edge loss, Charbonnier loss and Fundus quality perception loss & Fundus image restoration: Real Fundus dataset \\ \hline
    3D Transformer GAN \cite{3d_transformer_gan} & MICCAI'21 & Generator consisting of Encoder CNN, Transformer and Decoder CNN & CNN with four convolution blocks & Adversarial loss and L1 loss & PET reconstruction: PET data \\ \hline
    3D CVT-GAN \cite{3dcvtgan} & MICCAI'22 & 3D convolutional vision transformer (CVT) based encoder and 3D transposed CVT based decoder & Patch-based discriminator embedded with 3D CVT blocks & Adversarial loss and L1 loss & PET reconstruction: PET data \\ \hline
    Low-Light Transformer-GAN \cite{lowlight_transformer_gan} & IEEE-SPL'22 & Transformer using multi-head multi-covariance self-attention and Light feature-forward module structures & Convolutional discriminator & Adversarial loss, smooth L1 loss, perceptual loss, and multi-scale SSIM loss & Low-light enhancement: LOL and SICE \\ \hline
    LightingNet \cite{lightingnet} & IEEE-TCI'23 & Fusion of CNN-based encoder-decoder and ViT-based encoder-decoder & CNN-based discriminator & Adversarial loss, smooth L1 loss, perceptual loss, and multi-scale SSIM loss & Low-light enhancement: LOL, SICE, ExDARK, DICM, LIME, MEF, and NPE \\ \hline
    ICT \cite{ict} & ICCV'21 & Bi-directional transformer guided CNN & CNN-based discriminator & Adversarial loss and L1 loss & Image completion: FFHQ and Places2 \\ \hline
    BAT-Fill \cite{batfill} & ACMMM'21 & Bidirectional and autoregressive transformer + CNN-based texture generation & CNN-based discriminator & Adversarial loss, perceptual loss and reconstruction loss & Image inpainting: CelebA-HQ, Places2 and Paris StreetView \\ \hline
    \textit{T}-former \cite{tformer} & ACMMM'22 & U-shaped generator with transformer blocks & Patch GAN discriminator & Adversarial loss, style loss, reconstruction loss and perceptual loss & Image inpainting: CelebA-HQ, Places2 and Paris StreetView \\ \hline
    APT \cite{apt} & ACMMM'22 & Atrous pyramid transformer and dual spectral transform convolution & CNN-based discriminator & Adversarial loss, perceptual loss, style loss and L1 loss for masked and preserved regions & Image inpainting: CelebA-HQ, Places2 and Paris StreetView \\ \hline
    MAT \cite{mat} & CVPR'22 & A convolutional head, a mask-aware transformer body and a convolutional tail & Traditional discriminator & Adversarial loss, perceptual loss and R1 regularization & Large hole image inpainting: CelebA-HQ and Places365-Standard \\ \hline
    ZITS \cite{zits} & CVPR'22 & Transformer-based structure restorer + CNN-based structure feature encoding and texture restoration & PatchGAN discriminator & Adversarial loss, L1 loss over unmasked region, feature match loss and high receptive field perceptual loss & Image inpainting: Places2, ShanghaiTech, NYUDepthV2 and MatterPort3D \\ \hline
    HAN \cite{han} & ECCV'22 & Generator with CNN encoder, hourglass attention structure blocks and CNN decoder & PatchGAN discriminator with spectral norm & Adversarial loss, style loss, reconstruction loss and perceptual loss & Image inpainting: CelebA-HQ, Places2 and Paris StreetView \\ \hline
    SRInpainter \cite{srinpaintor} & IEEE-TCI'22 & Resolution progressive CNN encoder, hierarchical transformer and CNN decoder & SNPatchGAN discriminator & Adversarial loss and super-resolved L1 loss & Image inpainting: CelebA-HQ, Places2 and Paris StreetView \\ \hline
    NDMAL \cite{ndmal} & WACV'23 & Nested deformable attention layer mixed with convolution and de-convolution layers & PatchGAN discriminator & Adversarial loss, perceptual loss, edge loss and L1 loss & Image inpainting: CelebA-HQ and Places2 \\ \hline
    Hint \cite{hint} & WACV'22 & ViT-based generated hint converts outpaining to inpainting & Traditional discriminator & Adversarial loss, style loss, perceptual loss and L1 loss & Image outpainting: SUN and Beach \\ \hline
    ColorFormer \cite{color_former} & ECCV'22 & Generator using transformer-based encoder and a color memory decoder & PatchGAN discriminator & Adversarial loss, perceptual loss and content loss & Image colorization: ImageNet, COCO-Stuff and CelebA-HQ \\ \hline
    SGA \cite{sga} & ECCV'22 & Generator with stop-gradient attention module between encoder and decoder & Conditional CNN discriminator & Adversarial loss, perceptual loss, reconstruction loss and style loss & Image colorization: Anime portraits and Animal FacesHQ \\ \hline
    VTGAN \cite{vtgan} & ICCV'21 & CNN-based generator at different resolution & ViT for discriminator and classification at different resolution & Adversarial loss, mean square error, perceptual loss, embedding feature loss and cross-entropy loss & Retinal image synthesis and disease prediction using fundus and fluorescein angiogram images \\ \hline
    ResViT \cite{resvit} & IEEE-TMI'22 &  Transformer-based generator using aggregated residual transformer blocks & Conditional PatchGAN discriminator & Adversarial loss, reconstruction loss and pixel loss & Multimodal medical image synthesis: IXI brain MRI, BRATS and MRI-CT \\ \hline
\end{tabular}
\label{tab:image_translation}
\end{table*}

\section{Transformer-based GANs for Image-to-Image Translation}
The image generation mechanism generates the artificial sample from a random latent vector in a given data distribution. However, the image-to-image translation aims to transform the image from one domain to another domain. Hence, the generator network in GAN is modified such that it can take an image as the input and produces an image as the output. Mainly, encoder-decoder based architecture serves this purpose. Recently, several researchers have tried to investigate the Transformer-based GANs for the image-to-image translation tasks. We provide a summary of Transformer-based GAN models for image-to-image translation in Table \ref{tab:image_translation} in terms of generator model, discriminator model, objective function, applications and datasets.

\begin{table*}[!t]
    \caption{The results comparison of Transformer-based GANs for PET reconstruction on NC subjects and MCI subjects datasets. }
    \centering
    \begin{tabular}{p{0.35\columnwidth} p{0.15\columnwidth} p{0.08\columnwidth} p{0.08\columnwidth} p{0.12\columnwidth} p{0.08\columnwidth} p{0.08\columnwidth} p{0.12\columnwidth} p{0.09\columnwidth} p{0.09\columnwidth}}
    \hline
    & & \multicolumn{3}{c}{NC subjects} & \multicolumn{3}{c}{MCI subjects} & Params & GFLOPs \\
    Method & Venue & PSNR & SSIM & NMSE & PSNR & SSIM & NMSE \\ \hline
    3D Transformer-GAN \cite{3d_transformer_gan} & MICCAI'21 & \textit{24.818} & \textit{0.986} & \textit{0.0212} & \textit{25.249} & \textbf{0.987} & \textit{0.0231} & \textit{76M} & \textbf{20.78}\\
    3D CVT-GAN \cite{3dcvtgan} & MICCAI'22 & \textbf{24.893} & \textbf{0.987} & \textbf{0.0182} & \textbf{25.275} & \textbf{0.987} & \textbf{0.0208} & \textbf{16M} & \textit{23.80}\\
    \hline
    \end{tabular}
    \label{tab:results_pet_reconstruction}
\end{table*}

\subsection{Image Translation}
Image translation from a source domain to a target domain is an attractive application of GANs, such as Pix2pix \cite{pix2pix} and CycleGAN \cite{cyclegan} are very popular for such tasks.
Recently, Kim et al. discovered InstaFormer which exploits the transformer by integrating the global information with instance information for unpaired image translation \cite{instaformer}. The instance-level features are generated with the help of bounding box detection. InstaFormer utilizes the adaptive instance normalization and instance-level content contrastive loss to improve the local context encoding. The ViT encoder block \cite{vit} is used as a backbone in the generator network of InstaFormer. An impressive results are observed using Instaformer in terms of FID \& SSIM of 84.72 \& 0.872 on sunny$\rightarrow$night and 71.65 \& 0.818 on night$\rightarrow$sunny datasets, respectively.

The generator network of CycleGAN \cite{cyclegan} is replaced by a Vision Transformer (ViT) \cite{vit} for unsupervised image-to-image transformation in UVCGAN by Torbunov et al. \cite{uvcgan}. It is noticed from the experiments of UVCGAN that the self-supervised pre-training and gradient penalty are important for the improvements. UVCGAN reports the state-of-the-art FID scores of 79.0, 122.8, 9.6, 13.9, 14.4, and 13.6 on Selfie to Anime, Anime to Selfie, Male to Female, Female to Male, Remove Glasses, and Add Glasses datasets, respectively. 

Using ViT as a generator in GAN framework is computationally challenging. Zheng et al. \cite{ittr} proposed ITTR model for image-to-image translation using transformer for unpaired scenario. ITTR reduces the computational complexity by utilizing a dual pruned self-attention mechanism. In order to utilize global semantics, ITTR performs the mixing of tokens of varying receptive fields through hybrid perception blocks. The FID scores achieved by ITTR are 45.1, 68.6, 33.6, 73.4, 93.7, and 91.6 on Scene$\rightarrow$Semantic Map (Cityscapes), Cat$\rightarrow$Dog, Horse$\rightarrow$Zebra, Selfie$\rightarrow$Anime, Face$\rightarrow$Metface, and Female$\rightarrow$Cartoon datasets, respectively. Wang et al. \cite{piti} performs generative pretraining for the diverse downstream tasks to generate a highly semantic latent space as tokens using a transformer from text-image pairs. An adversarial diffusion upsampler is utilized for increasing the resolution of generated samples.

The image translation has been also a popular choice for medical applications using GANs. A Swin transformer GAN (MMTrans) is proposed in \cite{swin_transformer_GAN} for multi-modal medical image translation. The generator network of MMTrans is followed by a registration network. MMTrans uses a convolution-based discriminator network.

\subsection{Image Restoration}
Liang et al. proposed a Swin Transformer-based SwinIR model for image restoration \cite{swinir}. Basically, SwinIR utilizes several Swin Transformer layers together with a residual connection in a residual Swin Transformer block. SwinIR is tested for image super-resolution, image denoising and JPEG compression artifact restoration. It is reported that SwinIR leads to a reduction in the total number of parameters by up to 67\%, while achieves performance gain by up to 0.14$\sim$0.45dB.

Global-local stepwise GAN (GLSGN) is proposed by exploiting stepwise three local pathways and one global pathway based restoring strategy for high-resolution image restoration \cite{glsgn}. Interpathway consistency is applied for the mutual collaboration between four pathways of GLSGN. An impressive results are reported by GLSGN for high-resolution image dehazing, image deraining, and image reflection removal. However, the complexity of GLSGN is increased due to the uses of inputs at different resolution for different local pathways.

Transformer-based GAN (RFormer) is introduced by Deng et al. \cite{rformer} to clinical fundus images for restoration of the real degradation. In order to exploit the long-range dependencies and non-local self-similarity, a window-based self-attention block is utilized by RFormer. Moreover, a Transformer-based discriminator network is also used by RFormer. The experimental results with PSNR of 28.32 and SSIM of 0.873 are achieved for fundus image restoration by RFormer.

\subsection{Image Reconstruction}
Image reconstruction is very important w.r.t. biomedical applications, which can also be seen as image-to-image translation. Luo et al. \cite{3d_transformer_gan} proposed 3D Transformer-GAN for reconstruction of the standard-dose  positron emission tomography (SPET) image from the low-dose PET (LPET) image. The generator network in \cite{3d_transformer_gan} is developed as a CNN-based Encoder followed by a Transformer followed by a CNN-based Decoder. 3D Transformer-GAN shows the improved performance for the clinical PET reconstruction. A convolutional ViT based GAN, dubbed 3D CVT-GAN, is introduced by Zeng et al. for SPET reconstruction from LPET images \cite{3dcvtgan}. The encoder and decoder of 3D CVT-GAN use 3D CVT blocks for feature encoding and 3D transposed CVT (TCVT) blocks for SPET restoration, respectively. Table \ref{tab:results_pet_reconstruction} shows the results comparison between 3D Transformer-GAN \cite{3d_transformer_gan} and 3D CVT-GAN \cite{3dcvtgan} for PET reconstruction on normal control (NC) subjects and mild cognitive impairment (MCI) subjects datasets in terms of PSNR, SSIM, NMSE, parameters and GFLOPs. The higher values of PSNR \& SSIM and lower value of NMSE points out that 3D CVT-GAN is able to reconstruct the PET images with better quality. Moreover, 3D CVT-GAN is lighter than 3D Transformer-GAN in terms of the number of parameters. However, GFLOPs suggests that 3D Transformer-GAN leads to less number of operations as compared to 3D CVT-GAN.

A model-based adversarial transformer architecture (MoTran) is introduced for image reconstruction by Korkmaz et al. \cite{motran}. The generator of MoTran includes transformer and data-consistency blocks. MoTran reports better results than SLATER with best PSNR of 47.8 and SSIM of 0.992 on T2-weighted acquisitions of  IXI brain MRI dataset\footnote{http://brain-development.org/}. Zhao et al. \cite{swingan} proposed a SwinGAN which is a Swin Transformer-based GAN consisting of a frequency-domain generator and an image-domain generator for reconstruction of MRI images. SwinGAN achieves SSIM of 0.95 and PSNR of 32.96 on IXI brain dataset with 20\% undersampling rate, SSIM of 0.939 and PSNR of 34.50 on MRNet dataset of Knee \cite{mrnet}.

Chen et al. utilized a conditional GAN for cross-domain face synthesis and a mesh transformer for 3D face reconstruction \cite{transformer_3d_face_reconstruction}. Basically, conditional GAN translates the face images to a specific rendered style, which is exploited by transformer network to output 3D mesh vertices. The promising performance is reported in \cite{transformer_3d_face_reconstruction} for 3D face reconstruction using Transformer-based GAN.

\subsection{Low-light Image Enhancement}
Image-to-image translation is also useful for enhancing the low-light images. Wang et al. \cite{spgat} introduced a structural prior driven generative adversarial transformer (SPGAT) consisting of a structural prior estimator, a generator and two discriminators. The generator is a U-shaped transformer model. The efficacy of SPGAT is tested on both synthetic and real-world datasets for low-light image enhancement. In 2022, a Transformer-based GAN (Transformer-GAN) is used by Yang et al. \cite{lowlight_transformer_gan} for low-light image enhancement. In the first stage, the features are extracted by an iterative multi-branch network and image enhancement is performed in the second stage of image reconstruction. Basically, a ViT-based generator is combined with convolution-based discriminator in \cite{lowlight_transformer_gan}. Very recently, in 2023, Yang et al. proposed a LightingNet model for low-light image enhancement \cite{lightingnet}. LightingNet uses ViT-based low-light enhancement subnetwork along with a Res2Net-based complementary learning subnetwork. However, a CNN with 10 convolution layers is used as the discriminator network by LightingNet. The PSNR and SSIM scores of Transformer-based GANs are reported in Table \ref{tab:results_enhancement} for Low-light image enhancement over LOL \cite{lol} and SICE \cite{sice} datasets. Transformer-GAN \cite{lowlight_transformer_gan} achieves best PSNR and SSIM on LOL dataset. However, LightingNet \cite{lightingnet} performs better on the SICE dataset in terms of PSNR. It shows the suitability of transformers with GANs for image enhancement.

\begin{table}[!t]
    \caption{The results comparison of Transformer-based GANs for Low-light image enhancement over LOL \cite{lol} and SICE \cite{sice} datasets. }
    \centering
    \begin{tabular}{p{0.3\columnwidth} p{0.174\columnwidth} p{0.06\columnwidth} p{0.06\columnwidth} p{0.06\columnwidth} p{0.06\columnwidth} }
    \hline
    & & \multicolumn{2}{c}{LOL dataset} & \multicolumn{2}{c}{SICE dataset} \\
    Method & Venue & PSNR & SSIM & PSNR & SSIM \\ \hline
    SPGAT \cite{spgat} & arXiv'22 & 19.800 & 0.823 & - & - \\ 
    Transformer-GAN \cite{lowlight_transformer_gan} & IEEE-SPL'22 & \textbf{23.501} & \textbf{0.851} & \textit{21.902} & \textbf{0.878}\\
    LightingNet \cite{lightingnet} & IEEE-TCI'23 & \textit{22.735} & \textit{0.842} & \textbf{22.389} & \textit{0.801}\\
    \hline
    \end{tabular}
    \label{tab:results_enhancement}
\end{table}

\subsection{Image Super-resolution}
Transformer-based GANs have also shown its suitability for image super-resolution. SwinIR \cite{swinir}, proposed for image restoration, is also tested for image super-resolution on five benchmark datasets, resulting to outstanding performance. 
Kasem et al. utilized spatial transformer to develop a robust super-resolution GAN (RSR-GAN) \cite{spatial_transformer_gan}. Both the generator and discriminator networks of RSR-GAN use spatial transformer. The RSR-GAN shows promising performance on five benchmark super-resolution datasets. Du and Tian \cite{tgan} utilized the transformer and GAN (T-GAN model) for medical image super-resolution. The generator of T-GAN processes the input with two sub-networks, including a residual block based sub-network and a texture Transformer-based sub-network, and finally combines their output to generate the super-resolution image. The discriminator of T-GAN is a CNN-based network. The reported performance of T-GAN is PSNR of 34.92 \& SSIM of 0.94964 on Knee MRI images and PSNR of 34.69 \& SSIM of 0.9353 on Abdominal MRI images for super-resolution. Li et al. \cite{srinpaintor} proposed SRInpaintor by inheriting the characteristics of super-resolution using transformer for image inpainting. Coarse-to-fine information propagation and long-term relation encoding using hierarchical transformer are the key component of SRInpaintor. Very recently in 2023, Bao et al. proposed SCTANet model for face image super-resolution \cite{sctanet} which is a CNN-Transformer aggregation network by exploiting the spatial attention guidance. SCTANet utilizes a tail consisting of sub-pixel MLP-based upsampling module followed by a convolution layer. An outstanding performance is reported by SCTANet on CelebA and Helen face datasets for 4, 8 and 16 times super-resolution.

\begin{table}[!t]
    \caption{The results comparison of Transformer-based GANs for image inpainting over CelebA-HQ \cite{progressivegan} and Places2 \cite{places2} datasets. }
    \centering
    \begin{tabular}{p{0.06\columnwidth}p{0.217\columnwidth} p{0.174\columnwidth} p{0.05\columnwidth} p{0.06\columnwidth} p{0.05\columnwidth} p{0.06\columnwidth} }
    \hline
    & & & \multicolumn{2}{c}{CelebA-HQ} & \multicolumn{2}{c}{Places2} \\
    Mask & Method & Venue & SSIM & FID & SSIM & FID \\ \hline
    \multirow{4}{*}{30-40\%} & \textit{T}-former \cite{tformer} & ACMMM'22 &  \textbf{0.945} & \textbf{3.88} & 0.846 & \textit{26.56} \\ 
    & SRInpaintor \cite{srinpaintor} & IEEE-TCI'22 & 0.943 & 5.70 & \textit{0.862} & \textbf{11.24} \\
    & HAN \cite{han} & ECCV'22 & \textit{0.945} & \textit{3.93} & 0.839 & 28.85 \\
    & APT \cite{apt} & ACMMM'22 & - & - & \textbf{0.912} & - \\ 
    \hline

    \multirow{5}{*}{40-60\%} & ICT \cite{ict} & ICCV'21 & - & - & 0.739 & 34.21 \\
    & BAT-Fill \cite{batfill} & ACMMM'21 & 0.834 & \textbf{12.50} &  0.704 & \textbf{32.55} \\
    & PLSA \cite{plsa_vqgan} & HDIS'22 & - & - & \textbf{0.801} & \textit{33.14}\\
    & MAT \cite{mat} & CVPR'22 & \textit{0.847} & 13.12 & 0.726 & 35.81 \\
    & NDMAL \cite{ndmal} & WACV'23 & \textbf{0.858} & \textit{12.90} & \textit{0.776} & 37.89 \\
    \hline
    \end{tabular}
    \label{tab:results_inpainting}
\end{table}

\subsection{Image Inpainting and Outpainting}
Image inpainting/completion aims to fill the cracks and holes in the images. It is considered as an application of image-to-image translation. CNNs tend to suffer in understanding global structures for image completion due to some inherent characteristics (e.g., spatial invariant kernels and local inductive prior). Recently, transformers are also utilized for image inpainting as it can capture the global relationship.
In 2021, Wang et al. performed image inpainting using automatic consecutive context perceived transformer GAN (ACCP-GAN) \cite{accp_gan}. In the first stage, the broken areas are detected and repaired roughly using Convolution and gated Convolution-based sub-networks. In the second stage, the generated rough patches are refined by exploiting the serial perceive transformer which also exploits the information from neighboring images. ACCP-GAN is able to achieve 47.2252 FID on the N7 dataset for image inpainting. The transformer is combined with CNN in ICT \cite{ict} by first modeling the pluralistic coherent structures and some coarse textures using a transformer and then enhancing the local texture details using CNN. Impressive results on large-scale ImageNet dataset are obtained using ICT for image completion.

The recent methods exploit the masking with transformers \cite{mat}, \cite{zits}. 
A dynamic mask based attention module is utilized in Mask-Aware Transformer (MAT) \cite{mat} which aggregates non-local information only from partial valid tokens. MAT uses the transformer block between a convolutional head and a convolutional tail. MAT shows outstanding performance for high-resolution image inpainting. A Zero-initialized Residual Addition based Incremental Transformer Structure (ZITS) is proposed in \cite{zits} for image inpainting. ZITS exploits the orthogonal positional encoding in the masked regions. An impressive performance using ZITS is reported.
Hourglass attention network (HAN) and Laplace attention based transformer is proposed for image inpainting in \cite{han}. In 2022, \textit{T}-former is developed for image inpainting by Deng et al. \cite{tformer}. \textit{T}-former follows a linear attention based on Taylor expansion to reduce the computational complexity of self-attention for images. SRInpaintor performs the inpainting using transformer by exploiting the super-resolution characteristics \cite{srinpaintor}. 

Very recently, Phutke and Murala \cite{ndmal} proposed a nested deformable attention based transformer (NDMAL) for face image inpainting. The multi-head attention used in \cite{ndmal} consists of a deformable convolution leads to an efficient transformer model. Other Transformer-based GAN methods for image inpainting include atrous pyramid transformer and spectral convolution based model \cite{apt}, gated convolution and Transformer-based model \cite{generative_image_inpainting}, and visual transformers with multi-scale patch partitioning \cite{mspp}. The transformers have also been utilized for pluralistic image inpainting/completion \cite{ict}, \cite{plsa_vqgan}. The Swin Transformer-based models are used for image inpainting in \cite{sfiswin}, \cite{cswin_transformer}.

The results comparison in terms of SSIM and FID is performed in Table \ref{tab:results_inpainting} on CelebA-HQ \cite{progressivegan} and Places2 \cite{places2} datasets using different transformer driven GAN models for 30-40\% and 40-60\% mask. Higher SSIM and lower FID represent the better performance. It is noticed that \textit{T}-former \cite{tformer} performs best on CelebA-HQ dataset for 30-40\% mask. However, NDMAL \cite{ndmal} shows highest SSIM for 40-60\% mask on CelebA-HQ dataset. On Places2 dataset, APT \cite{apt} and PLSA \cite{plsa_vqgan} lead to highest SSIM for 30-40\% and 40-60\% mask, respectively. It is also observed that BAT-Fill \cite{batfill} is a better model to maintain a lower FID score across both the datasets.

Similar to image inpaining, image outpainiting is also performed as image-to-image translation where the outer area is reconstructed. Transformer-based GANs are also investigated for image outpaining, such as U-Transformer \cite{utransformer_outpainting} and ViT-based Hint \cite{hint}. U-Transformer exploits Swin transformer blocks for U-shaped encoder-decoder \cite{utransformer_outpainting}. Hint method attaches the representative hint at the boundaries and converts outpainting problem into inpainting problem \cite{hint}. The representative hint is generated using ViT on different patches in \cite{hint}.

\subsection{Image Colorization}
Image colorization is performed as image-to-image translation. Recently, few Transformer-based GANs have shown promising performance for image colorization. A color memory powered hybrid-attention transformer (ColorFormer) is introduced by Ji et al. \cite{color_former} for image colorization. The encoder of ColorFormer is a global-local hybrid attention based transformer. The decoder of ColorFormer utilizes a color memory decoder for image-adaptive queries through semantic-color mapping. A stop-gradient attention (SGA) mechanism is utilized by Li et al. \cite{sga} which removes the conflicting gradient and becomes better suitable for reference-based line-art colorization. An impressive results are obtained for four datasets. SGA is also utilized for Anime line drawing colorization \cite{sga_ext}. A dual decoder based DDColor model consisting of a transformer-based color decoder and a multi-scale image decoder is investigated for image colorization in \cite{ddcolor}. Plausible results of image colorization are obtained using DDColor model.
A cycle swin transformer GAN (CSTGAN) is proposed in \cite{cstgan} for colorizing the infrared images in an unpaired scenario. CSTGAN combines two swin transformer blocks with one convolution block in a module and connect such modules with skip connections in the generator network.

\subsection{Medical Image Synthesis}
Li et al. \cite{slmtnet} performed cross-modality MR image synthesis using a self-supervised learning based multi-scale transformer network (SLMT-Net). A pretrained ViT encoder using edge information is followed by a multi-scale transformer U-Net to produce the image in target modality. Better PSNR scores are obtained using the SLMT-Net model for MR image synthesis.
A transformer generator is used to enhance the input image with global information followed by a CNN generator in TCGAN model \cite{tcgan}. Basically, TCGAN takes a positron emission tomography (PET) image as the input and synthesizes the corresponding computer tomography (CT) image as the output. In an another Residual Transformer Conditional GAN (RTCGAN) work, MR image is synthesized into the corresponding CT image by Zhao et al. \cite{rtcgan}. RTCGAN encodes the local texture information using CNN and global correlation using Transformer. RTCGAN achieves an impressive SSIM of 0.9105. A cycle-consistent Transformer (CyTran) model utilizing convolutional transformer block is used to synthesize the contrast CT images from the corresponding non-contrast CT images \cite{cytran}. 

The fundus retinal images are translated into corresponding angiogram images in VTGAN \cite{vtgan}, where two vision transformers are utilized as the discriminators-cum-classifiers for coarse and fine images, respectively. Basically, the vision transformer in VTGAN discriminates between the real and fake samples and at the same time outputs the normal vs. abnormal class labels. Different source-target modality configurations are combined into a ResViT model in \cite{resvit}. ResViT utilizes an Encoder-Decoder based generator which uses a residual convolutional and Transformer-based building blocks. ResViT enjoys the precision of convolution operators along with the contextual sensitivity of vision transformers. ResViT is demonstrated for the synthesis of missing frames in multi-contrast MRI, and CT images from MRI. A multi-view transformer-based generator by exploiting cross-view attention mechanism is introduced in \cite{mvt} for cardiac cine MRI reconstruction. The multi-view transformer model \cite{mvt} is able to focus on the important regions in different views for the reconstruction.

\begin{table*}[!t]
\caption{A summary of Transformer-based GANs for video applications.}
\centering
\begin{tabular}{p{0.18\columnwidth}|p{0.15\columnwidth}|p{0.5\columnwidth}|p{0.25\columnwidth}|p{0.36\columnwidth}|p{0.33\columnwidth}}
    \hline
    \textbf{Model} & \textbf{Venue} & \textbf{Generator} & \textbf{Discriminator} & \textbf{Objective Function} & \textbf{Application \& Datasets} \\
    \hline
    TATS \cite{tats} & ECCV'22 & Generator with time-agnostic 3D VQGAN and time-sensitive Trasformer & Two discriminators: a spatial discriminator and a temporal discriminator & VQGAN: Adversarial loss, matching loss, reconstruction loss, codebook loss, commit loss, Transformer: Negative log-likelihood & Video generation: UCF-101, Sky Time-lapse and Taichi-HD \\ \hline
    MAGViT \cite{magvit} & arXiv'22 & Generator consisting of the 3D-VQ Encoder, Bidirectional Transformer and 3D-VQ Decoder & StyleGAN-based 3D discriminator & GAN loss, image perceptual loss, LeCam regularization, reconstruction loss, refine loss, and masking loss &  Video generation: Multi-task with 10 tasks including prediction, interpolation, inpainiting, and outpainting \\ \hline
    ActFormer \cite{actformer} & arXiv'22 & Generator having action-conditioned interaction and temporal transformers & Graph Convolutional Network  & Conditional Wasserstein GAN loss & Motion generation: NTU-13, NTU RGB+D 120, BABEL, and GTA Combat \\ \hline
    FuseFormer \cite{fuseformer} & ICCV'21 & A sub-token fusion enabled Transformer with a soft split and composition method with CNN encoder and decoder & CNN-based video discriminator & Adversarial loss and reconstruction loss & Video inpainting: DAVIS and YouTube-VOS \\ \hline
    Style Transformer \cite{aast} & IEEE-TMM'22 & Generator with a deep encoder, axial attention block, transformer, and decoder & Temporal PatchGAN-based discriminator & Adversarial loss, L1 loss, and reconstruction loss & Video inpainting: DAVIS and YouTube-VOS \\ \hline
    DeViT \cite{devit} & ACMMM'22 & Generator with Encoder, Patch-based deformed Transformer, and decoder & Temporal PatchGAN-based discriminator & Adversarial loss and reconstruction loss on hole and valid pixels & Video inpainting: DAVIS and YouTube-VOS \\ \hline
    FGT \cite{fgt} & ECCV'22 & Generator with flow-guided content propagation, spatial and temporal Transformers & Temporal PatchGAN-based discriminator & Adversarial loss and reconstruction loss & Video inpainting: DAVIS and YouTube-VOS \\ \hline
    FGT++ \cite{fgt_extended} & arXiv'23 & FGT with flow-guided feature integration and flow-guided feature propagation modules & Temporal PatchGAN-based discriminator & Adversarial loss, spatial domain reconstruction loss and amplitude loss & Video inpainting: DAVIS and YouTube-VOS \\ \hline
    CT-D2GAN \cite{ct-d2gan} & ACMMM'21 & Convolutional transformer with encoder, temporal self-attention module and decoder & Two discriminators: 2D Conv and 3D Conv-based discriminators & Adversarial loss and pixel-wise L1 loss & Video anomaly detection: UCSD Ped2, CUHK Avenue, and ShanghaiTech Campus dataset \\ \hline
    Trajectory Transformer \cite{trajectory_prediction} & IJIS'21 & Transformer with multi-head convolutional self‐attention & Discriminator with a decoder and prediction module & Adversarial loss and L2 loss & Trajectory prediction: ETH and UCY datasets \\ \hline
    Bidirectional Transformer GAN \cite{btgan} & ACM-TMCCA'23 & Transformer-based motion generator for forward and backward processing & Frame-based and Sequence-based discriminators & Adversarial loss, inverse loss, and soft dynamic time warping (Soft-DTW) loss & Human motion prediction: public Human3.6M dataset \\ \hline
    MaskViT \cite{maskvit} & ICLR'23 & Generator with VQGAN and Bidirectional window transformer for variable percentage masked tokens prediction & VQ-GAN discriminator & Adversarial loss, perceptual loss and reconstruction loss & Video prediction: BAIR, RoboNet and KITTI datasets \\ \hline
    Recurrent Transformer Network \cite{rtn} & CVPR'22 & A bi-directional RNN architecture having temporal aggregation module in masked encoder and flow features and spatial restoration transformer followed by Conv layers & Temporal PatchGAN discriminator & Spatial-temporal adversarial loss, L1 loss and perceptual loss & Video colorization: DAVIS and REDS dataset\\
    \hline
\end{tabular}
\label{tab:video_applications}
\end{table*}

\subsection{Other Image-to-Image Translation Applications}
Transformer-based GANs are also utilized for different other image translation applications. A Transformer-based CycleGAN is proposed in \cite{transformer_liver_segmentation} for liver tumor segmentation. A CASTformer is proposed in \cite{castformer} for 2D medical image segmentation, which is a class-aware transformer for learning the important object semantic regions. Trans-CycleGAN is developed in \cite{transcyclegan} by utilizing the Transformer-based generator and discriminator networks for image style transfer. 

A symmetric and semantic aware transformer (SSAT) is exploited in \cite{ssat} for makeup transfer and removal by learning the semantic correspondences. Pose guided human image synthesis (PGHIS) is performed in \cite{pghis} by exploiting a transformer module between the encoder and decoder networks. The synthesis using PGHIS is performed on the decoupled human body parts (e.g., face, hair, feet, hands, etc.). Pedestrian synthesis is performed using a pose and color-gamut guided GAN (PC-GAN) in \cite{pc-gan}. The generator in the PC-GAN consists of a local displacement estimator, a color-gamut transformer, and a pose transporter. PC-GAN is used to improve the performance of person re-identification by augmenting the training set.

In \cite{asset}, a Transformer-based autoregressive semantic scene editing (ASSET) method is developed for high-resolution images based on the user's defined semantic map edits. ASSET uses dense attention in the transformer at lower image resolutions and sparsifies the attention matrix at high resolutions. Xu et al. proposed a Transformer-based GAN, named TransEditor, for facial editing \cite{transeditor}. TransEditor exploits dual-space editing and inversion strategy to enhance the interaction in a dual-space GAN to provide additional editing flexibility. The results indicate that  TransEditor is effective for highly controllable facial editing.

In 2023, a cycle transformer GAN (CTrGAN) is introduced by Mahpod et al. \cite{ctrgan} for Gait transfer having Transformer-based generator and discriminator networks. The source image is first converted into a dense pose which is then translated into the most natural dense pose of the target using CTrGAN. Finally, the image of the target is synthesized by rendering the generated pose. CTrGAN shows promising results for Gait transfer.
FusionGAN performs the depth estimation using Transformer-based GAN as a multimodal image translation \cite{fusiongan}. FusionGAN utilizes the late fusion of the features of the transformer block with the sparse depth map and RGB image, which is followed by residual in residual dense block and convolution layer to produce the dense depth map.

\subsection{Summary}
Following is the summary drawn from the Transformer-based GANs for image-to-image translation:
\begin{itemize}
    \item It is noticed that both global and local contexts play an important roles for image-to-image translation. Hence, the majority of the methods exploit both the Transformer and Convolution in the generator network.
    \item The generator generally consists of Encoder, Transformer and Decoder modules. Vision transformer based architectures are heavily utilized. Some methods also try to modify the self-attention module with task specific information.
    \item CNN-based discriminators are mostly exploited, such as PatchGAN discriminator. Some regularizers are also used by few models, such as gradient penalty and spectral normalization.
    \item The L1 loss, perceptual loss, reconstruction loss, and style loss are commonly used in the objective function along with adversarial loss.
    \item Transformer-based GANs have shown state-of-the-art results for several image-to-image translation applications, such as image translation, image reconstruction, image restoration, image enhancement, image super-resolution, image inpainting \& outpainting, image colorization, medical image synthesis, image segmentation, human and pose synthesis, image editing and depth prediction. 
\end{itemize}

\section{Transformer-based GANs for Video Applications}
In recent years, several attempts have been made to utilize the Transformer-based GANs for different video processing applications. A summary on Transformer-based GANs is presented in Table \ref{tab:video_applications} for different video applications in terms of generator network, discriminator network, objective function, application and datasets.

\subsection{Video Synthesis}
In 2022, Ge et al. \cite{tats} synthesized the longer videos of thousands of frames using Time-Agnostic VQGAN and Time-Sensitive Transformer (TATS). The hierarchical transformer in TATS helps to capture longer temporal dependencies. The experiments on benchmark UCF-101, Sky Time-lapse, and Taichi-HD datasets confirm the suitability of TATS \cite{tats} for generating the longer videos.
Yu et al. proposed a Masked Generative Video Transformer (MAGViT) \cite{magvit} by first 3D tokenizing the video and then applying an embedding method to model the masked video tokens. MAGViT is able to perform for diverse video generation from different visual appearances.
In 2022, Xu et al. proposed ActFormer model to generate the action-conditioned 3D human motion from a latent vector for the frames of single-person as well as multi-person interactions \cite{actformer}.

\subsection{Video Translation}
\subsubsection{Video Inpainting}
Video inpainting is a very important application of video translation. In 2021, Liu et al. proposed a Transformer-based FuseFormer model for video inpainting \cite{fuseformer}. FuseFormer exploits soft split and composition method to perform sub-token fusion. In 2022, a deformed vision transformer (DeViT) is proposed for video inpainting \cite{devit}. Deformed patch homography based patch alignment and mask pruning based patch attention are exploited by DeViT. The attention to spatial-temporal tokens is obtained by a spatial-temporal weighting adaptor module in \cite{devit}. A generator, consisting of deep encoder, axial attention block, transformer, and decoder, is utilized in Axial Attention-based Style Transformer (AAST) \cite{aast} for video inpainting. The transformer in \cite{aast} exploits the high-frequency patch information between temporal and spatial features. Flow guided transformer (FGT) is also utilized for video inpainting by integrating the completed optical flow into the transformers \cite{fgt}, \cite{fgt_extended}. 

The results comparison is performed in Table \ref{tab:results_video_inpainting} for video inpainting using transformer-based GAN models on YouTube-VOS \cite{youtube-vos} and DAVIS \cite{davis} datasets in terms of PSNR and SSIM. FGT++* with flow-guided content propagation \cite{fgt_extended} performs best for video inpainting. It is observed that the utilization of flow information with the transformer is very beneficial for video inpainting.

\subsubsection{Other Video Translation Applications}
Feng et al. proposed a CT-D2GAN model for future frame synthesis \cite{ct-d2gan}, consisting of a convolutional transformer generator, a 2D convolutional discriminator and a 3D convolutional discriminator. The generator in CT-D2GAN is made with a convolutional encoder followed by a temporal self-attention block followed by a convolutional decoder. Very recently, MaskViT is introduced for future frame prediction in videos \cite{maskvit}. MaskViT first converts the video frames into tokens using VQGAN and then randomly mask some tokens of the future frames and performs pre-training of a Bidirectional Window Transformer for masked token prediction. At the inference time, frames are predicted in an iterative refinement fashion with incrementally decreasing masking ratio. A recurrent transformer network (RTN) is introduced in \cite{rtn} for the restoration of degraded old films. RTN exploits the useful information from adjacent frames for ensuring temporal coherency and restoring the challenging artifacts.
A bidirectional transformer GAN (BTGAN) is proposed in \cite{btgan} for human motion generator. BTGAN follows the CycleGAN framework with motion sequences as input to generator and predicted motion sequences as output. A new soft dynamic time warping (Soft-DTW) loss is utilized by BTGAN for training the generator.
Transformer-based GANs are also utilized for pedestrian trajectory prediction in \cite{trajectory_prediction} by learning the pedestrian distribution to generate more reasonable future trajectories.

\begin{table}[!t]
    \caption{The results comparison of Transformer-based GANs for video inpainting on YouTube-VOS \cite{youtube-vos} and DAVIS \cite{davis} datasets. The use of flow-guided content propagation is represented by *. }
    \centering
    \begin{tabular}{p{0.23\columnwidth} p{0.2\columnwidth} p{0.07\columnwidth} p{0.07\columnwidth} p{0.07\columnwidth} p{0.07\columnwidth}}
    \hline
    & & \multicolumn{2}{c}{Youtube-VOS} & \multicolumn{2}{c}{DAVIS} \\
    Method & Venue & PSNR & SSIM & PSNR & SSIM \\ \hline
    FuseFormer \cite{fuseformer} & ICCV'21 & 33.16 & 0.967 & 32.54 & 0.970 \\
    DeViT \cite{devit} & ACMMM'22 & 33.42 & 0.973 & 32.43 & 0.972\\
    AAST \cite{aast} & IEEE-TMM'22 & 33.23 & 0.967 & 32.71 & 0.972 \\
    FGT \cite{fgt} & ECCV'22 & 34.04 & 0.971 & 32.60 & 0.965 \\
    FGT* \cite{fgt} & ECCV'22 & 34.53 & \textit{0.976} & \textit{33.41} & \textit{0.974} \\
    FGT++ \cite{fgt_extended} & arXiv'23 & \textit{35.02} & \textit{0.976} & 33.18 & 0.971 \\
    FGT++* \cite{fgt_extended} & arXiv'23 & \textbf{35.36} & \textbf{0.978} & \textbf{33.72} & \textbf{0.976} \\
    \hline
    \end{tabular}
    \label{tab:results_video_inpainting}
\end{table}

\subsection{Summary}
Following is the summary on Transformer-based GANs for video applications:
\begin{itemize}
    \item Most of the models convert video into tokens using VQGAN-based Encoder and then apply the spatial and/or temporal transformers.
    \item The flow information is also exploited and utilized at different levels in the generator of some models.
    \item Temporal PatchGAN discriminator is utilized by a majority of the models. Convolution discriminators are also used by some models.
    \item The objective function usually contains adversarial loss, reconstruction loss, L1 loss and perceptual loss.
    \item Transformer-based GANs have shown promising performance for different video applications, such as video generation, video inpainting, video prediction, video anomaly detection and video colorization.
    \item It is noticed that the flow-guided feature integration and propagation is very important with transformers for video inpainting. 
\end{itemize}

\section{Transformer-based GANs for Miscelleneous Applications}
Apart from image and video synthesis, Transformer-based GANs have also been exploited for other applications, such as text-to-image generation, hyperspectral image classification, document image enhancement, etc.

\subsection{Text-to-Image Generation}
Transformer-based GANs have been utilized for text-to-image generation \cite{transformer_attngan}, \cite{dsegan}, \cite{layout_vqgan}, \cite{muse}. Naveen et al. performed text-to-image generation experiments with AttnGAN using different Transformer models such as BERT, GPT2, and XLNet \cite{transformer_attngan}. An improvement of 49.9\% in FID is observed in \cite{transformer_attngan}. Huang et al. proposed a dynamical semantic evolution GAN (DSE-GAN) \cite{dsegan} for text-to-image generation. The generator of DSE-GAN is composed of a single adversarial multi-stage structure having a series of grid-Transformer-based generative blocks which are weighted by a series of DSE modules at different stages. An object-guided joint decoding transformer is introduced by Wu et al. \cite{layout_vqgan} for generating the image and the corresponding layout from the text. The layout is encoded and decoded using a Layout-VQGAN to extract additional useful features for the synthesis of the complex scenes. In 2023, a masked generative transformer, named Muse, is investigated for text-to-image generation \cite{muse}. Muse is trained on a masked modeling task to predict randomly masked image tokens using pre-trained large language model (LLM). Muse shows very promising performance on the benchmark datasets, including FID of 7.88 on zero-shot COCO evaluation using a 3B parameter model and FID of 6.06 on CC3M dataset using a 900M parameter model.

\subsection{Hyperspectral Image Classification}
The ViT-based generator, discriminator, and classifier networks are utilized in HyperViTGAN model \cite{hypervitgan} for hyperspectral image (HSI) classification. The classifier network classifies the input HSI patch into one of the class categories, whereas the discriminator network classifies the input HSI patch into real or fake category. The improved results on three benchmark HSI datasets are obtained by HyperViTGAN. Similarly, a transformer-based generator network is used for synthesizing the class-specific HSI patches in \cite{transgan_hsi} to alleviate the data imbalance problem in HSI classification.

\subsection{Other Applications}
Kodym and Hradiš \cite{tg2} generated high-quality text line image from the corresponding degraded low-quality text line image and the corresponding transcription as an input. The low-quality image is translated into the high-quality image using an Encoder-Decoder network. The transcription information is aligned with the output of Encoder using transformer consisting of multi-head attention modules. A promising performance is recorded for the document image restoration tasks, including inpainting, debinarization, and deblurring.
A transformer model is used for anomaly detection in \cite{transformer_based_gan} by exploiting the holistic features of different classes. In order to encode global semantic information, self-attention modules are used in generator of GAN in \cite{transformer_based_gan}. 
A cross-modal Transformer-based GAN (CT-GAN) is utilized in \cite{ct-gan} for synthesizing the multi-modal brain connectivity. Basically, CT-GAN fuses the structural information and functional information from different imaging modalities. Two decoders are used to extract the functional connectivity and structural connectivity from the fused connectivity. Two discriminators are also employed in \cite{ct-gan} corresponding to two decoders.

\subsection{Summary}
Overall, it is noticed that Transformer-based GANs are very successful for different image applications, including text-to-image generation, hyperspectral image classification, document analysis, anomaly detection, etc. The use of transformer-based GANs is still limited to various computer vision applications leaving to a huge scope for future research in this domain.

\section{Conclusion and Future Directions}
\label{conclusion}
\subsection{Conclusion and Trend}
This paper presents a comprehensive survey of Transformer-based Generative Adversarial Networks (GANs) for computer vision applications. As most of the Transformer-based GANS are very recent, the progress and advancements in the past few years, i.e., 2020-2023 are presented. The survey is conducted for different type of applications, including image generation, image-to-image translation, video generation, video translation and other computer vision applications. The detailed discussion is provided with further application specific categorization, such as translation, restoration, reconstruction, inpainting, super-resolution, etc. A summary of different Transformer-based GAN models is presented in terms of generator architecture, discriminator architecture, objective functions, applications and datasets. The performance analysis using the state-of-the-art Transformer-based GAN models are also presented in terms of FID/PSNR/SSIM for different applications, such as image generation, high-resolution image generation, PET image reconstruction, low-light image enhancement, image inpainting and video inpainting. 

The research trend in image and video synthesis points out that the Transformer-based GANs are the latest driving factor in the progress of computer vision methods and applications. The generator with recently developed Transformer, Vision Transformer and Swin Transformer models has shown the superior performance for image and video synthesis. The utilization of both local context and global context using the Convolution and Transformer networks is heavily utilized by several models. For translation task, U-shaped generator with Convolution-based Encoder followed by Transformer followed by Convolution-based Decoder is very well exploited. The StyleGAN with Swin Transformer has shown outstanding performance for image generation. For video translation task, the flow information is exploited with Transformers. Mostly, Transformer-based GANs rely on CNN-based discriminator. However, few models also tried to exploit the Transformers for discriminator. The utilization of suitable loss functions with adversarial loss is also the recent trend in order to improve the synthesis quality, perceptual quality, etc. Other trends include prediction of masked tokens, designing of novel attention module, utilization of cross-attention and convolutional transformers, etc.

\subsection{Future Directions}
The future works in Transformer-based GANs include development of generator and discriminator architectures for different applications. There is a huge scope to find better ways to fuse the Convolution and Transformer features in order to exploit the local and global information. The modification in self-attention module with task specific characteristics is also a potential future direction. Other future direction involves the advancements in Vision and Swin Transformers by exploiting the power of state-of-the-art GAN models, such as StyleGAN, etc. The masking-based pre-training of transformer model can also be further explored. A major future aspect is to utilize the Transformer-based GANs for different applications of image and video processing in computer vision. Exploitation of flow information with Transformers can be explored further for video applications. The pre-training of Transformer-based GANs with diverse applications using large-scale datasets, development of light-weight models, utilization of self-supervision, etc. are the further scope to progress. The development of better loss functions and the identification of suitable combination of losses in objective function can also be performed in the future. The exploration to find the better hyperparameter settings to stabilize the training and to increase the generalization capability of the Transformer-based GAN models.

{\small
\bibliographystyle{IEEEtran}
\bibliography{Ref}
}

\begin{IEEEbiography}[{\includegraphics[width=1in,height=1.25in,clip,keepaspectratio]{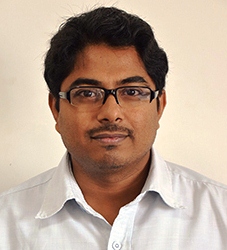}}]{Shiv Ram Dubey} is with the Indian Institute of Information Technology (IIIT), Allahabad since July 2021, where he is currently the Assistant Professor of Information Technology. He was with IIIT Sri City as Assistant Professor from Dec 2016 to July 2021 and Research Scientist from June 2016 to Dec 2016. He received the PhD degree from IIIT Allahabad in 2016. Before that, from 2012 to 2013, he was a Project Officer at Indian Institute of Technology (IIT), Madras. He was a recipient of several awards including the Best PhD Award in PhD Symposium at IEEE-CICT2017, Early Career Research Award from SERB, Govt. Of India and NVIDIA GPU Grant Award Twice from NVIDIA. Dr. Dubey is serving as the Treasurer of IEEE Signal Processing Society Uttar Pradesh Chapter.
His research interest includes Computer Vision and Deep Learning.
\end{IEEEbiography}

\begin{IEEEbiography}[{\includegraphics[width=1in,height=1.25in,clip,keepaspectratio]{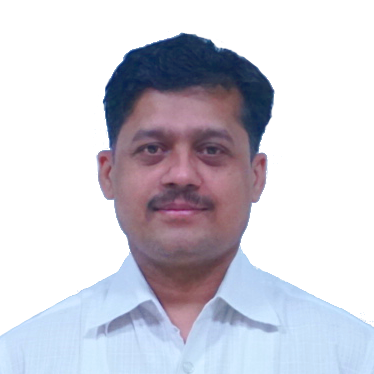}}]{Satish Kumar Singh}
is with the Indian Institute of Information Technology Allahabad, as an Associate Professor from 2013 and heading the Computer Vision and Biometrics Lab (CVBL). Earlier, he served at Jaypee University of Engineering and Technology Guna, India from 2005 to 2012. His areas of interest include Image Processing, Computer Vision, Biometrics, Deep Learning, and Pattern Recognition. Dr. Singh is proactively offering his volunteer services to IEEE for the last many years in various capacities. He is the senior member of IEEE. Presently Dr. Singh is the Section Chair IEEE Uttar Pradesh Section (2021-2022) and a member of IEEE India Council (2021). He also served as the Vice-Chair, Operations, Outreach and Strategic Planning of IEEE India Council (2020) \& Vice-Chair IEEE Uttar Pradesh Section (2019 \& 2020). Prior to that Dr. Singh was Secretary, IEEE UP Section (2017 \& 2018), Treasurer, IEEE UP Section (2016 \& 2017), Joint Secretary, IEEE UP Section (2015), Convener Web and Newsletters Committee (2014 \& 2015). 
Dr. Singh is also the technical committee affiliate of IEEE SPS IVMSP and MMSP and presently the Chair of IEEE Signal Processing Society Chapter of Uttar Pradesh Section. 
\end{IEEEbiography}

\end{document}